\documentclass[11pt,a4paper]{article}
\usepackage[hyperref]{acl2019}
\usepackage[utf8]{inputenc}
\usepackage{times}
\usepackage{latexsym}
\usepackage{booktabs}
\usepackage{tikz}
\usepackage{adjustbox}
\usepackage{multirow}

\usepackage{url}
\usepackage{todonotes}

\newcommand{\minornotesoff}{\long\gdef\minornote##1{}}
\newcommand{\minornoteson}{\long\gdef\minornote##1{\par\noindent\fbox{\parbox
{0.45\textwidth}{{\large MINOR NOTE.} \small\scshape ##1}}\\[0.3ex]}}
\minornoteson
\minornotesoff

\aclfinalcopy 

\newcommand{\cls}{\texttt{\small [cls]}}

\title{How Language-Neutral is Multilingual BERT?}

\author{Jindřich Libovický$^1$ \and Rudolf Rosa$^2$ \and Alexander Fraser$^1$ \\
\\
$^1$Center for Information and Language Processing, LMU Munich, Germany \\
$^2$Faculty of Mathematics and Physics, Charles University, Prague, Czech Republic \\
\texttt{\{libovicky, fraser\}@cis.lmu.de rosa@ufal.mff.cuni.cz}
    }

\date{}

\begin{document}

\maketitle

\begin{abstract}
Multilingual BERT (mBERT) provides sentence representations for 104 languages,
which are useful for many multi-lingual tasks.
Previous work probed the cross-linguality of mBERT using zero-shot transfer
learning on morphological and syntactic tasks.
We instead focus on the semantic properties of mBERT\@.
We show that mBERT representations can be split into a language-specific
component and a language-neutral component, and that the language-neutral
component is sufficiently general in terms of modeling semantics to allow
high-accuracy word-alignment and sentence retrieval but is not yet good enough
for the more difficult task of MT quality estimation.
Our work presents interesting challenges which must be solved to build better
language-neutral representations, particularly for tasks requiring linguistic
transfer of semantics.
\end{abstract}

\section{Introduction}




Multilingual BERT (mBERT\@; \citealt{devlin2019bert}) is gaining popularity as
a contextual representation for various multilingual tasks, such as dependency
parsing \citep{kondratyuk2019udify,wang2019crosslingual}, cross-lingual natural
language inference (XNLI) or named-entity recognition (NER)
\citep{pires2019multilingual,wu2019beto,kudugunta2019investigating}.

\citet{pires2019multilingual} present an exploratory paper showing that mBERT
can be used cross-lingually for zero-shot transfer in morphological and
syntactic tasks, at least for typologically similar languages. They also study
an interesting semantic task, sentence-retrieval, with promising initial
results. Their work leaves many open questions in terms of how good the
cross-lingual mBERT representation is for semantics, motivating our work.

In this paper, we directly assess the semantic cross-lingual properties of
mBERT\@. To avoid methodological issues with zero-shot transfer (possible
language overfitting, hyper-parameter tuning), we selected tasks that only
involve a direct comparison of the representations: cross-lingual sentence
retrieval, word alignment, and machine translation quality estimation (MT QE).
Additionally, we explore how the language is represented in the embeddings by
training language identification classifiers and assessing how the
representation similarity corresponds to phylogenetic language families.

\minornote{AF: this next paragraph doesn't seem right -- shouldn't it introduce
adversarial too? I'd spend a paragraph talking about ways we try to strengthen
the language-neutral part, using the language-neutral terminology introduced in
the abstract. Then I'd spend another paragraph talking about what happens
(including maybe the projection result).}

Our results show that the mBERT representations, even after language-agnostic
fine-tuning, are not very language-neutral. However, the identity of the
language can be approximated as a constant shift in the representation space.
An even higher language-neutrality can still be achieved by a linear projection
fitted on a small amount of parallel data.

Finally, we present attempts to strengthen the language-neutral component via
fine-tuning: first, for multi-lingual syntactic and morphological analysis;
second, towards language identity removal via a adversarial classifier.

\section{Related Work}

Since the publication of mBERT \citep{devlin2019bert}, many positive
experimental results were published.

\citet{wang2019crosslingual} reached impressive results in zero-shot dependency
parsing. However, the representation used for the parser was a bilingual
projection of the contextual embeddings based on word-alignment trained on
parallel data.

\citet{pires2019multilingual} recently examined the cross-lingual properties of
mBERT on zero-shot NER and  part-of-speech (POS) tagging but the success of
zero-shot transfer strongly depends on how typologically similar the languages
are. Similarly, \citet{wu2019beto} trained good multilingual models for POS
tagging, NER, and XNLI, but struggled to achieve good results in the zero-shot
setup.
%
%

\minornote{AF: Can we say here: ``In contrast, we study semantic tasks'' or
something similar? There needs to be a contrast with our work.}

\citet{pires2019multilingual} assessed mBERT on cross-lingual sentence
retrieval between three language pairs. They observed that if they subtract the
average difference between the embeddings from the target language
representation, the retrieval accuracy significantly increases. We
systematically study this idea in the later sections.

Many experiments show
\citep{wu2019beto,kudugunta2019investigating,kondratyuk2019udify} that
downstream task models can extract relevant features from the multilingual
representations. But these results do not directly show language-neutrality,
i.e., to what extent are similar phenomena are represented similarly across
languages. The models can obtain the task-specific information based on the
knowledge of the language, which (as we show later) can be easily identified.
Our choice of evaluation tasks eliminates this risk by directly comparing the
representations.
%
%
Limited success in zero-shot setups and the need for explicit bilingual
projection in order to work well
\citep{pires2019multilingual,wu2019beto,ronnqvist2019multilingual} also shows
limited language neutrality of mBERT\@.

\section{Centering mBERT Representations}

Following \citet{pires2019multilingual}, we hypothesize that a sentence
representation in mBERT is composed of a language-specific component, which
identifies the language of the sentence, and a language-neutral component,
which captures the meaning of the sentence in a language-independent way. We
assume that the language-specific component is similar across all sentences in
the language.

We thus try to remove the language-specific information from the
representations by centering the representations of sentences in each language
so that their average lies at the origin of the vector space. We do this by
estimating the language centroid as the mean of the mBERT representations for a
set of sentences in that language and subtracting the language centroid from
the contextual embeddings.

We then analyze the semantic properties of both the original and the centered
representations using a range of probing tasks. For all tasks, we test all
layers of the model. For tasks utilizing a single-vector sentence
representation, we test both the vector corresponding to the \cls{} token and
mean-pooled states.

\section{Probing Tasks}

We employ five probing tasks to evaluate the language neutrality of the
representations.

\paragraph{Language Identification.} 
With a representation that captures all phenomena in a language-neutral way, it
should be difficult to determine what language the sentence is written in.
Unlike other tasks, language identification does require fitting a classifier.
We train a linear classifier on top of a sentence representation to try to
classify the language of the sentence.

\paragraph{Language Similarity.} 
Experiments with POS tagging \citep{pires2019multilingual} suggest that similar
languages tend to get similar representations on average. We quantify that
observation by measuring how languages tend to cluster by the language families
using V-measure over hierarchical clustering of the language centeroid
\citep{rosenberg2007vmeasure}.

\paragraph{Parallel Sentence Retrieval.} 

For each sentence in a multi-parallel corpus, we compute the cosine distance of
its representation with representations of all sentences on the parallel side
of the corpus and select the sentence with the smallest distance.

Besides the plain and centered \cls{} and mean-pooled representations, we
evaluate explicit projection into the ``English space''. For each language, we
fit a linear regression projecting the representations into English
representation space using a small set of parallel sentences.

\paragraph{Word Alignment.} 
While sentence retrieval could be done with keyword spotting, computing
bilingual alignment requires resolving detailed correspondence on the word
level.

We find the word alignment as a minimum weighted edge cover of a bipartite
graph. The graph connects the tokens of the sentences in the two languages and
edges between them are weighted with the cosine distance of the token
representation.  Tokens that get split into multiple subwords are represented
using the average of the embeddings of the subwords.
\minornote{AF: I doubt this next sentence can be understood at this point in
the paper. Move it later maybe? Or explain it in detail?}
Note that this algorithm is invariant to representation centering which would
only change the edge weights by a constant offset.

We evaluate the alignment using the F$_1$ score over both sure and possible
alignment links in a manually aligned gold standard.

\paragraph{MT Quality Estimation.} 
MT QE
assesses the quality of an MT system output without having access to
a reference translation.

The standard evaluation metric is the correlation with the
\minornote{AF: IMPORTANT - this says HTER, but aren't you using TER?!?}
Human-targeted Translation Error Rate which is the number of edit operations a
human translator would need to do to correct the system output. This is a more
challenging task than the two previous ones because it requires capturing more
fine-grained differences in meaning.

We evaluate how cosine distance of the representation of the source sentence
and of the MT output reflects the translation quality.
\minornote{AF: please check the next two sentences}
In addition to plain and centered representations, we also test trained
bilingual projection, and a fully supervised regression trained on training
data.

\section{Experimental Setup}


We use a pre-trained mBERT model that was made public with the BERT
release\footnote{https://github.com/google-research/bert}. The model dimension
is 768, hidden layer dimension 3072, self-attention uses 12 heads, the model
has 12 layers. It uses a vocabulary of 120k wordpieces that is shared for all
languages.

To train the language identification classifier, for each of the BERT languages
we randomly selected 110k sentences of at least 20 characters from Wikipedia,
and keep 5k for validation and 5k for testing for each language. The training
data are also used for estimating the language centroids.

For parallel sentence retrieval, we use a multi-parallel corpus of test data
from the WMT14 evaluation campaign \citep{bojar2014findings} with 3,000
sentences in Czech, English, French, German, Hindi, and Russian.
The linear projection experiment uses the WMT14 development data.
%

We use manually annotated word alignment datasets to evaluate word alignment
between English on one side and Czech (2.5k sent.;
\citealp{marecek2016alignment}), Swedish (192 sent.;
\citealp{holmqvist2011gold}), German (508 sent.), French (447 sent.;
\citealp{och2000improved}) and Romanian (248 sent.;
\citealp{mihalcea2003evaluation}) on the other side.
\minornote{AF: this should say something about FastAlign being an oracle
experiment, since it has access to additional information.}
We compare the results with FastAlign \citep{dyer2013simple} that was provided
with 1M additional parallel sentences from ParaCrawl \citep{espla2019paracrawl}
in addition to the test data.

For MT QE, we use English-German data provided for the WMT19 QE Shared Task
\citep{fonseca2019findings} consisting training and test data with source
senteces, their automatic translations, and manually corrections.

\section{Results}



\begin{table}[t]

	\centering

    \begin{tabular}{lccc}
        \toprule
        & mBERT & UDify & lng-free \\ \midrule

        \cls{}           & .935 & .938 & .796 \\
        \cls, cent.      & .867 & .851 & .337 \\ \midrule
        mean-pool        & .919 & .896 & .230 \\
        mean-pool, cent. & .285 & .243 & .247 \\ \bottomrule

    \end{tabular}
	\caption{Accuracy of language identification, values from the best-scoring
	layers.}\label{tab:lngid}

\end{table}

\paragraph{Language Identification.}

Table~\ref{tab:lngid} shows that centering the sentence representations
considerably decreases the accuracy of language identification, especially in
the case of mean-pooled embeddings. This indicates that the proposed centering
procedure does indeed remove the language-specific information to a great
extent.

\begin{figure}[t]

    \centering
    \begin{adjustbox}{clip,trim=0cm 1.0cm 0cm 0.8cm}
    \includegraphics[width=.9\columnwidth]{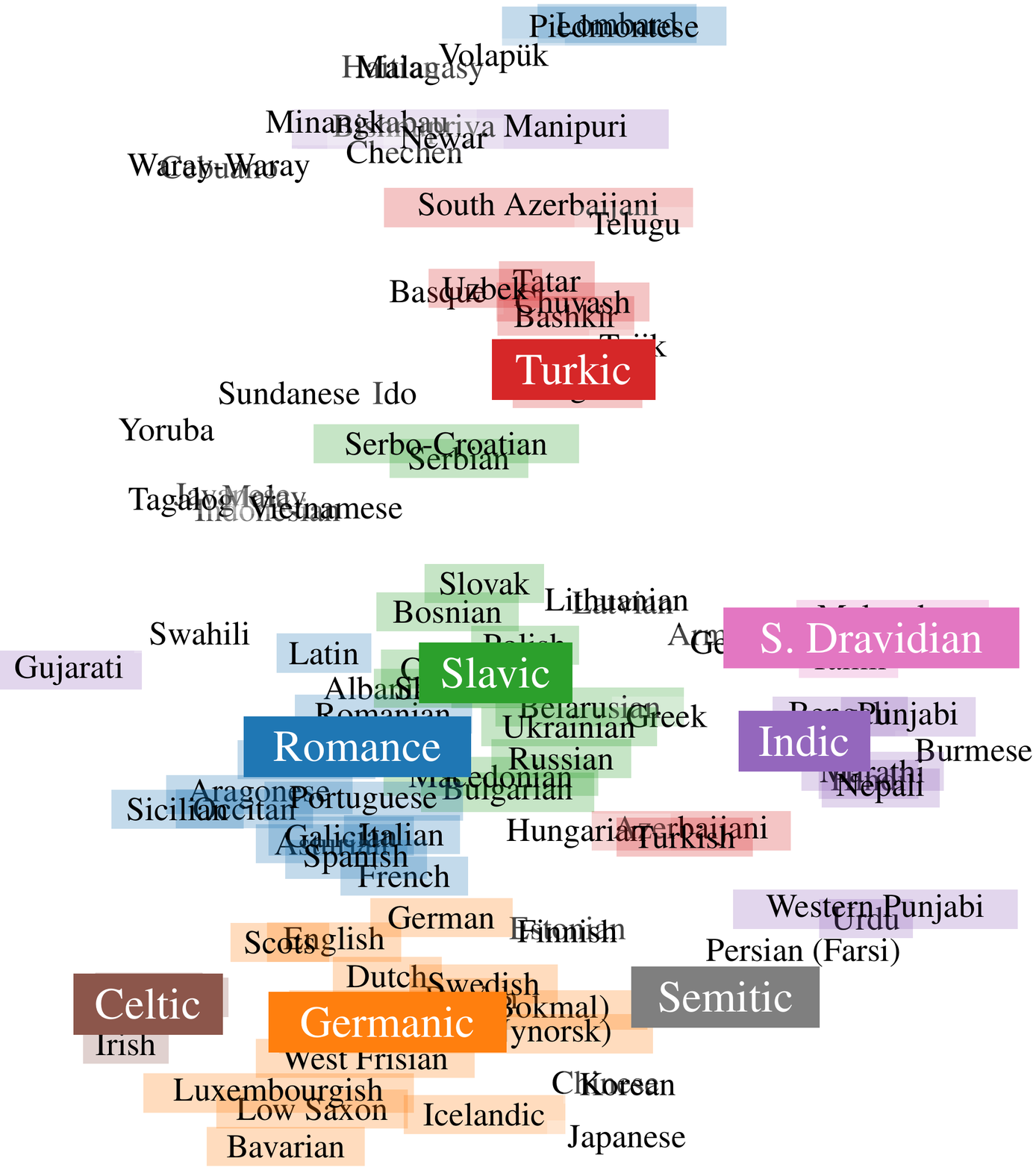}
    \end{adjustbox}
    \caption{Language centroids of the mean-pooled representations from the 8th
    layer of cased mBERT on a tSNE plot with highlighted language
    families.}\label{fig:families}

\end{figure}

\begin{table}[t]

    \begin{tabular}{ccccc}
        \toprule
        cased & uncased & UDify & lng-free & random \\ \midrule
        82.42 & 82.09   & 80.03 & 80.59    & 62.14 \\ \bottomrule
    \end{tabular}
    \caption{V-Measure for hierarchical clustering of language centroids
    and grouping languages into genealogical families for families with at
    least three languages covered by mBERT.}\label{tab:families}

\end{table}

\paragraph{Language Similarity.}

Figure~\ref{fig:families} is a tSNE plot \citep{maaten2008visualizing} of the
language centroids, showing that the similarity of the centroids tends to
correspond to the similarity of the languages. Table~\ref{tab:families}
confirms that the hierarchical clustering of the language centroids mostly
corresponds to the language families.

\begin{table}[t]

    \begin{tabular}{lccc}
        \toprule
        & mBERT & UDify & lng-free \\ \midrule

        \cls{}           & .639 & .462 & .549 \\
        \cls, cent.      & .684 & .660 & .686 \\
        \cls, proj.      & .915 & .933 & .697 \\ \midrule
        mean-pool        & .776 & .314 & .755 \\
        mean-pool, cent. & .838 & .564 & .828 \\
        mean-pool, proj. & .983 & .906 & .983 \\
        \bottomrule

    \end{tabular}
    \caption{Average accuracy for sentence retrieval over all 30 language
        pairs.}\label{tab:retrieval}

\end{table}

\begin{figure}[t]

    \centering
    \begin{adjustbox}{clip,trim=0cm 2.9cm 0cm 2.8cm}
    \includegraphics[width=.9\columnwidth]{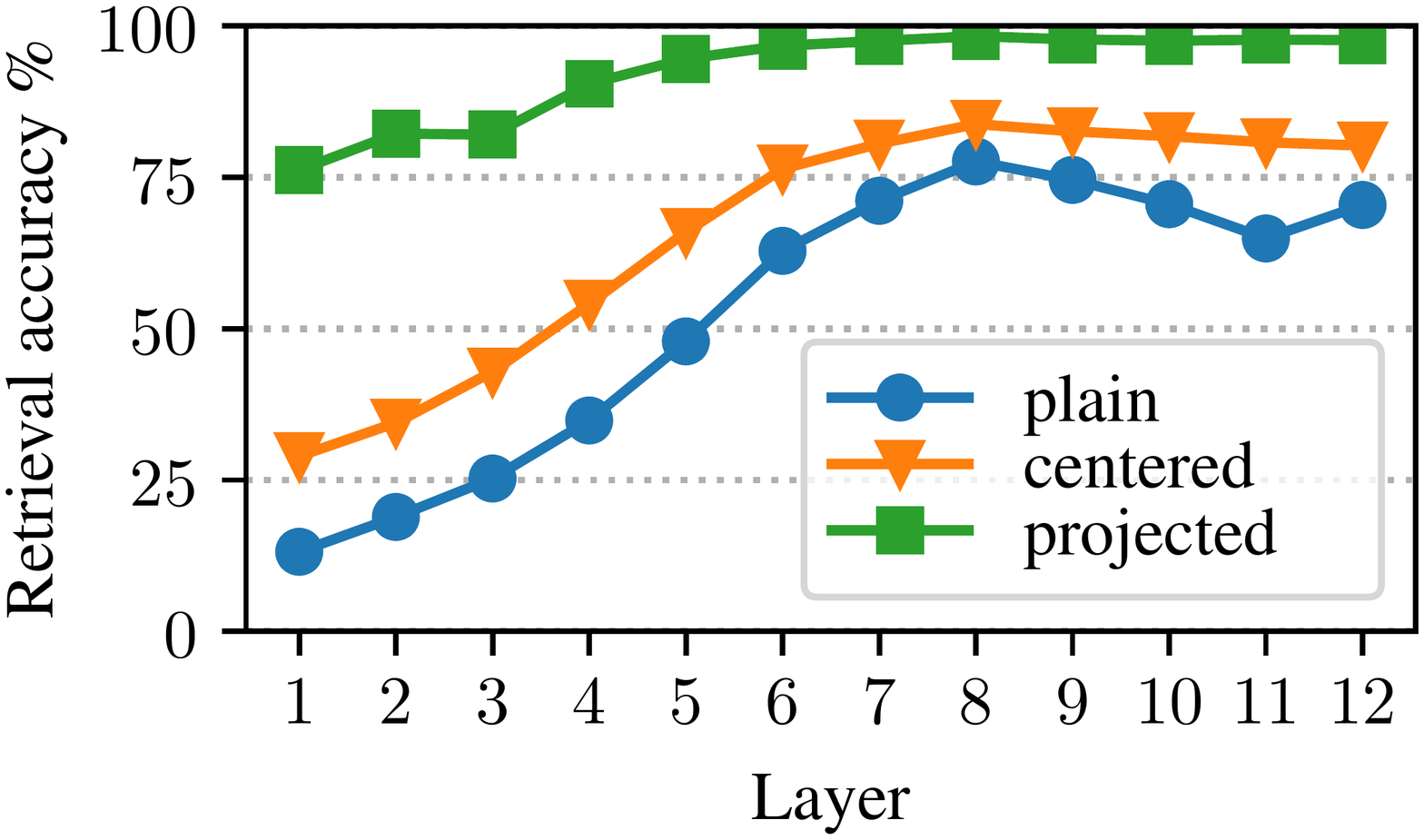}
    \end{adjustbox}
    \caption{Accuracy of sentence retrieval for mean-pooled contextual
    embeddings from BERT layers.}\label{fig:retrieval_layers}

\end{figure}

\paragraph{Parallel Sentence Retrieval.}

Results in Table~\ref{tab:retrieval} reveal that the representation centering
dramatically improves the retrieval accuracy, showing that it makes the
representations more language-neutral. However, an explicitly learned
projection of the representations leads to a much greater improvement, reaching
a close-to-perfect accuracy, even though the projection was fitted on
relatively small parallel data. The accuracy is higher for mean-pooled states
than for the \cls{} embedding and varies according to the layer of mBERT used
(see Figure~\ref{fig:retrieval_layers}).



\begin{table}[t]

    \begin{tabular}{lcccc}
        \toprule
        en- & \scalebox{0.8}[1.0]{FastAlign} & mBERT & UDify & lng-free \\ \midrule

        cs & .692 & .738 & .708 & .744 \\
        sv & .438 & .478 & .459 & .468 \\
        de & .471 & .767 & .731 & .768 \\
        fr & .583 & .612 & .581 & .607 \\
        ro & .690 & .703 & .696 & .704 \\

        \bottomrule

    \end{tabular}
    \caption{Maximum F$_1$ score for word alignment across
    layers compared with FastAlign baseline.}\label{tab:alignment}

\end{table}

\paragraph{Word Alignment.}

\minornote{AF: this footnote is hard to understand, but probably you should
leave it. I'd suggest additionally saying ``details are omitted due to
space.''}

Table~\ref{tab:alignment} shows that word-alignment based on mBERT
representations surpasses the outputs of the standard FastAlign tool even if it
was provided large parallel corpus. This suggests that word-level semantics are
well captured by mBERT contextual embeddings. For this task, learning an
explicit projection had a negligible effect on the performance.%
\footnote{We used an expectation-maximization approach that alternately aligned
the words and learned a linear projection between the representations. This
algorithm only brings a negligible improvement of .005 F$_1$ points.}




\begin{table}[t]
    \centering
    \begin{tabular}{lcccccc}\toprule
        \multirow{2}{*}{BERT} &
        cente- &
        glob. & \multicolumn{3}{c}{supervised} \\ \cmidrule(lr){4-6}
        & red & proj. & src & MT & both \\
         \midrule

        cased    &  .005 & .163 & .362 & .352 & .419 \\
        uncased  &  .027 & .204 & .367 & .390 & .425 \\
        UDify    &  .039 & .167 & .368 & .375 & .413 \\
        lng-free &  .026 & .136 & .349 & .343 & .411 \\

         \bottomrule
    \end{tabular}
    \caption{Correlation of estimated MT quality with HTER for
    English-to-German translation on WMT19 data.}\label{tab:qe}

\end{table}

\paragraph{MT Quality Estimation.}

\minornote{AF: this result on MT QE without using the hypothesis is really
interesting. Maybe highlight this more? Or is this maybe well known, I haven't
seen people doing this before, but I am not that familiar with what people have
been doing lately.}

Qualitative results of MT QE are tabulated in Table~\ref{tab:qe}.
%
%
Unlike sentence retrieval, QE is more sensitive to subtle differences between
sentences. Measuring the distance of the non-centered sentence vectors does not
correlate with translation quality at all. Centering or explicit projection
only leads to a mild correlation, much lower than a supervisedly trained
regression;%
\footnote{Supervised regression using either only the source or only MT output
also shows a respectable correlation, which implies that structural features of
the sentences are more useful than the comparison of the source sentence with
MT output.}%
and even better performance is possible \citep{fonseca2019findings}.
%
%
The results show that the linear projection between the representations only
captures a rough semantic correspondence, which does not seem to be sufficient
for QE\@, where the most indicative feature appears to be sentence complexity.

\section{Fine-tuning mBERT}\label{sec:finetuning}

We also considered model fine-tuning towards stronger language neutrality. We
evaluate two fine-tuned versions of mBERT\@: \emph{UDify}, tuned for a
multi-lingual dependency parser, and \emph{lng-free}, tuned to jettison the
language-specific information from the representations.

\subsection{UDify}

The UDify model \citep{kondratyuk2019udify} uses mBERT to train a single model
for dependency parsing and morphological analysis of 75 languages. During the
parser training, mBERT is fine-tuned, which improves the parser accuracy.
Results on zero-shot parsing suggest that the fine-tuning leads to more
cross-lingual representations with respect to morphology and syntax.

However, our analyses show that fine-tuning mBERT for multilingual dependency
parsing does not remove the language identity information from the
representations and actually makes the representations less semantically
cross-lingual.




\begin{figure}[t]

    \centering
    \begin{adjustbox}{clip,trim=0cm 2.9cm 0cm 2.8cm}
    \includegraphics[width=.9\columnwidth]{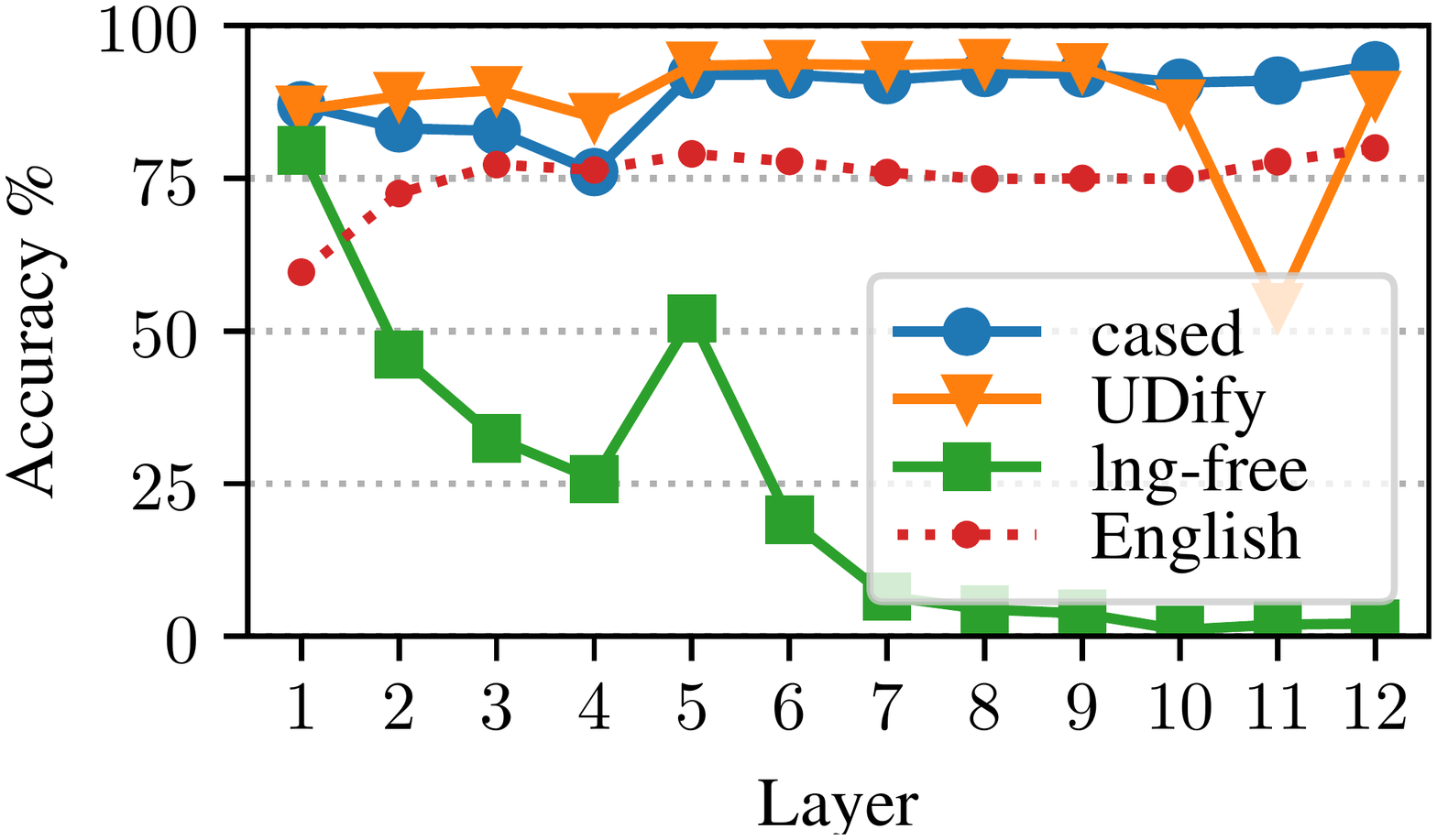}
    \end{adjustbox}
%
    \caption{Language ID accuracy for different layers of
    mBERT.}\label{fig:lngid_layers}

\end{figure}

\subsection{lng-free}

\minornote{AF: is this briefly discussed somewhere previously? If not, it might come as a surprise here? I am not sure having this in a subsection at the end is the right place for it.}

In this experiment, we try to make the representations more language-neutral by
removing the language identity from the model using an adversarial approach.
We continue training mBERT in a multi-task learning setup with the masked LM
objective with the same sampling procedure \citep{devlin2019bert} jointly with
adversarial language ID classifiers \citep{elazar2018adversarial}. For each
layer, we train one classifier for the \cls{} token and one for the mean-pooled
hidden states with the gradient reversal layer \citep{ganin2015unsupervised}
between mBERT and the classifier.

The results reveal that the adversarial removal of language information
succeeds in dramatically decreasing the accuracy of the language identification
classifier; the effect is strongest in deeper layers for which the standard
mBERT tend to perform better (see Figure~\ref{fig:lngid_layers}). However,
other tasks%
%
%
are not affected by the adversarial fine-tuning.

\section{Conclusions}

Using a set of semantically oriented tasks that require explicit semantic
cross-lingual representations, we showed that mBERT contextual embeddings do
not represent similar semantic phenomena similarly and therefore they are not
directly usable for zero-shot cross-lingual tasks.

Contextual embeddings of mBERT capture similarities between languages and
cluster the languages by their families. Neither cross-lingual fine-tuning nor
adversarial language identity removal breaks this property. A part of language
information is encoded by the position in the embedding space, thus a certain
degree of cross-linguality can be achieved by centering the representations for
each language. Exploiting this property allows a good cross-lingual sentence
retrieval performance and bilingual word alignment (which is invariant to the
shift). A good cross-lingual representation can be achieved by fitting a
supervised projection on a small parallel corpus.

\minornote{AF: but bad MT QE?}

\minornote{AF: what did we fail to show?}

\minornote{AF: what will future language-neutral representations be able to do
that mBERT (and other things we studied) can't do? Or equally: what are the
challenges we are making to future creators of language-neutral
representations?}


\bibliography{references}
\bibliographystyle{acl_natbib}

\end{document}